\documentclass{article}

\usepackage{arxiv}
\usepackage{subfigure}
\usepackage{color}
\usepackage[utf8]{inputenc} 
\usepackage[T1]{fontenc}    
\usepackage{hyperref}       
\usepackage{url}            
\usepackage{booktabs}       
\usepackage{amsfonts}       
\usepackage{nicefrac}       
\usepackage{microtype}      
\usepackage{lipsum}
\usepackage{graphicx}
\graphicspath{ {./images/} }

\title{SeMemNN: A Semantic Matrix-Based Memory Neural Network for Text Classification}

\author{
 Changzeng Fu$^{*1,2}$, Chaoran Liu$^{2}$, Carlos Toshinori Ishi$^{2}$,Yuichiro Yoshikawa$^{1}$, Hiroshi Ishiguro$^{*1,2}$\\
  $^{1}$Graduate School of Engineering Science, Osaka University, Japan\\
  $^{2}$Advanced Telecommunications Research Institute International, Japan\\
  \texttt{changzeng.fu@irl.sys.es.osaka-u.ac.jp} \\
 }

\begin{document}
\maketitle
\begin{abstract}
Text categorization is the task of assigning labels to documents written in a natural language, and it has numerous real-world applications including sentiment analysis as well as traditional topic assignment tasks. In this paper, we propose 5 different configurations for the semantic matrix-based memory neural network with end-to-end learning manner and evaluate our proposed method on two corpora of news articles (AG news, Sogou news). The best performance of our proposed method outperforms the baseline VDCNN models on the text classification task and gives a faster speed for learning semantics. Moreover, we also evaluate our model on small scale datasets. The results show that our proposed method can still achieve better results in comparison to VDCNN on the small scale dataset. This paper is to appear in the Proceedings of the 2020 IEEE 14th International Conference on Semantic Computing (ICSC 2020), San Diego, California, 2020. 
\end{abstract}


\section{Introduction}
Text classification is a classic task in the natural language processing domain, in which one requires a machine learning model to learn semantics from text samples in order to achieve a good performance on assigning predefined categories to free-text documents.

In text classification research field, the state-of-the-art methods have long been linear predictors (e.g. linear kernel SVM \cite{joachims1998text,lewis2004rcv1}) with either bag-of-word or bag-of-n-gram vectors as input. More recently, it has become more common to use a deep neural network for text classification. These previous studies using the deep neural network can be mainly divided into the CNN-based genre (e.g. CNN \cite{johnson2014effective}, char-CNN \cite{zhang2015character}, VDCNN \cite{conneau2016very}, DPCNN \cite{johnson2017deep}), transformer genre (e.g. BERT \cite{devlin2018bert}, XLNet \cite{yang2019xlnet}), and fine-tuning genre (e.g. ULMFiT \cite{rother2018ulmfit}). However, these methods highly rely on large-scale data sets and long-term training dependencies. Even though the ULMFiT model can reach state-of-the-art performance with only 100 labeled training samples, it still needs a large scale of samples for pre-training. This consumption limits the generalizability of models in new fields/tasks. In these cases, we considered that it is better to find a method to realize fast training and small sample dependency, so that meta-learning hereby comes to our mind. 

Meta-learning \cite{mitchell1980need}, or learning to learn, is the science of systematically observing how the learning approaches perform on learning tasks, and then learning from this experience or meta-data. It is not only dramatically speed up and improve the design of machine learning pipelines or neural architectures, but also allows us to replace hand-engineered algorithms with novel approaches learned in a data-driven way \cite{vilalta2002perspective}. Santoro et al. \cite{santoro2016one} proposed a method which is quite similar to the mechanism of meta-learning. They used an external memory matrix for augmenting memory of a neural network, which achieved amazing performance on one-shot learning for the image classification task. They utilized this external memory matrix to record and update memories, then verbatim retrieving these memories in the prediction. Although this mechanism can record the memory well, for the text classification task, the model not only needs to record the knowledge but also needs to understand the knowledge because of the variety of word combinations. 

In this work, we proposed a neural network based on Memory Network \cite{santoro2016one}  (MemNN) to update a semantic matrix that compresses the learned semantics and constructs a text understanding mechanism, we name it as semantic memory neural network (SeMemNN). The proposed model was evaluated on AG News and Sogou News and their subset by shrinking the amount of training data (see Table 1). 
The major contributions of this work is summarized as follows.

1) Classifying text based on SeMemNN in an end-to-end (E2E) manner.

2) Proposing 5 different configurations for SeMemNN \footnote[1]{Code for this work \url{https://github.com/CZFuChason/SeMemNN}}.

3) Outperforming VDCNN on the text classification task.

4) The proposed method gives a faster speed for learning semantics.

5) The proposed method is able to learn semantics from small scale datasets.

The paper is structured in the following way: In Section 2, we describe the details of our proposed method. The experiments along with the evaluation results are described in Section 3. Some discussions and future works are presented in Section 4. Section 5 is a brief summary of our work.

\section{Model Architecture}
\begin{figure*}
    \centering
    \includegraphics[width=1\textwidth]{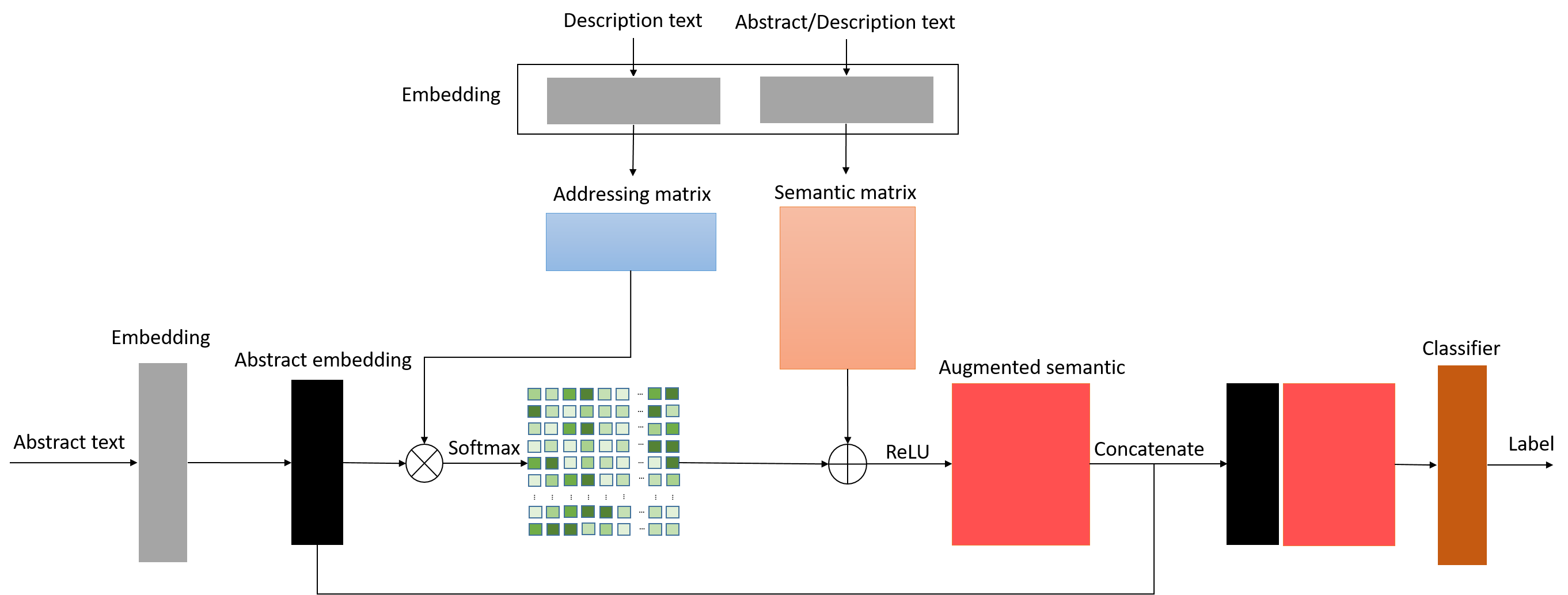}
    \caption{The architecture of proposed model}
    \label{fig:1}
\end{figure*}
Before diving into the details of SeMemNN, let me briefly recall what MemNN is and how do we humans do in the text classification task. It has been confirmed that MemNN and its derived models (e.g. end-to-end Memory Network \cite{sukhbaatar2015end}, Key-Value Memory Network \cite{miller2016key}, etc.) outperform some conventional question and answer (QA) system that are based on LSTM or knowledge base in QA task. The core of inference for MemNN lies in the $O$ and $R$ (see equations (1) and (2)). Assuming that the input of MemNN is to be a sentence, the $O$ produces output features by finding k supporting memories given the feature $x$, which is similar to addressing function. In equation (1), $s_{O}$ is a function that scores the pair of sentences and $m_{i}$ is the later supporting memory given in the previous iteration. The candidate supporting memory $m_{i}$ is scored with respect to the original input and the previous supporting memory. The final output is the input to the module $R$, which produces a textual response $r$. As equation (2) shows, the $w$ is the set for all words in the vocabulary, and $s_{R}$ is a function that scores the match. The simplest responses are to directly return the matching memory. The scoring function is shown as equation (3), where $U$ is a $n \times D$ matrix, $D$ is the number of features and $n$ is the embedding dimension. The role of $\phi_{x}$ and $\phi_{y}$ is to map the original text to the $D$-dimensional feature space. 

\begin{equation}
o_{n}=O_{n}(x,m)=\mathop{arg\max}_{i=1,...,N} \ s_{O}(x,m_{i})  
\end{equation}

\begin{equation}
r=O_{n}(x,m)=\mathop{arg\max}_{w \epsilon W} \ s_{R}([x,m_{o}],w)  
\end{equation}

\begin{equation}
s(x, y)=\phi_{x}(x)^{\top}U^{\top}U\phi_{y}(y)
\end{equation}

As for how we humans do in task classification, when we read and try to understand a piece of text, we always firstly attempt to extract key information (abstract) from the contents, then look for the corresponding prior knowledge in our memory to confirm which field the text belongs to. If this text is new for us, we are about to record and compress it to our knowledge base.

By referring to the cues of humans' behaviors and using external storage to augment the memory, we proposed a neural network with a semantic matrix to augment learned semantics. Figure 1 depicts the architecture of our proposed model. We defined two external matrices, one is an addressing matrix, another one is a semantics matrix. The whole text classification procedures involve three steps:

\paragraph{\textbf{Addressing}} As equation 4 shows, we use the information contained in the abstract to multiply with the addressing matrix to get the address tensor. The $\phi$ are feature maps of dimension $D$, $P$ is a $d\times D$ matrix. $Des_{i}$ is the description text, $Ab_{i}$ indicates the abstract of each text data. And $Softmax(z_{i})=e^{z_{i}}/\sum_{j}e^{z_{j}}$.
\begin{equation}
Addr=Softmax(P\phi_{Des}(Des_{i})\cdot P\phi_{Ab}(Ab_{i}))
\end{equation}

\paragraph{\textbf{Reading Semantics}} In the reading step, the semantics are read by augmenting the parameter's value by adding address tensor with semantic matrix and go through a ReLU activation for selecting the corresponding semantics with the highest contribution value according to the obtained matching scores. The new learned semantic will be pushed into the semantics matrix constantly by backpropagation in each training episode. The $\varphi$ in equation (5) is a semantic feature map of dimension $M$, $Z$ is a $m\times M$ matrix.  $C_{i}$ and $S$ indicate the contents of each text data and parameters in the semantic matrix respectively. $f$ is a ReLU activation that we used for emphasizing the most recently used semantic location and ignoring the least used memory location.

\begin{equation}
S =  Z\varphi_S({C_{i}})
\end{equation}
\begin{equation}
O = f(Addr + S)
\end{equation}

\paragraph{\textbf{Classification}}
The augmented semantic tensor is concatenated with the abstract embedding before passing to the classifier. Regarding the classifier, there are many types of neural layers that can be chosen. Since semantic is a sort of information lying on the time sequence of textual contents, modeling these temporal contexts effectively is a key to the text classification task. The most common approach is using a recurrent neural network (RNN) and its variants. Long short term memory (LSTM) has been used in several works to improve the model's ability to catch the long-term dependency in a time series \cite{lee2015high,le2017discretized}. Moreover, For catching the interesting time steps in a sequence, the attention mechanism is also added to the LSTMs and showing its effectiveness \cite{zhang2019attention}. Therefore, we adopt a recurrent neural network based model with attention mechanism. We then compared the performance of using different classifiers in the testing experiment.

\section{Experiment}
\begin{table}[b]
\centering
\caption{Large-scale text classification data sets}
\label{tab:1}       
\begin{tabular}{lllll}
\hline\noalign{\smallskip}
  Data set & Train & Test & classes & Task \\
\noalign{\smallskip}\hline\noalign{\smallskip}
AG news & 120k & 7.6k & 4 & English news categorization\\
AG news & 5k & 7.6k & 4 & English news categorization\\
Sogou news & 450k & 60k & 5 & Chinese news categorization\\
Sogou news & 10k & 60k & 5 & Chinese news categorization\\
\noalign{\smallskip}\hline
\end{tabular}
\end{table}
In this section, we demonstrate the performance of our models for text classification with different configurations, namely three different classifiers and two different sources to construct the semantics matrix. These three configurations are double-layer LSTM, one-layer bi-directional LSTM, one-layer bi-directional LSTM with self-attention. we note these different configurations as L-SeMemNN, B-SeMemNN, and SAB-SeMemNN respectively. Two different sources for constructing a knowledge matrix are the abstract (abs) and descriptions/contents (ct) of the news. 

\subsection{Datasets}

We present our results on two freely available large scale datasets introduced by Zhang et al. \cite{zhang2015character}. We also shrink the scale of training samples by randomly picking up the same amount of data in each category (see Table 1).

\paragraph{AG News} The AG News corpus consists of news articles from the AG’s corpus of news articles on the web pertaining to the 4 largest classes, which are \textit{Work}, \textit{Sports}, \textit{Business}, \textit{Sci/Tech}. The data set contains 30,000 training samples for each class, 1,900 samples for each class for testing. In this data set, there are three columns which are \emph{label}, \emph{title}, \emph{description}, we treat \emph{title} as abstract input and \emph{description} as contents input.

\paragraph{Sogou News} A Chinese news data set. This data set is a combination of the SogouCA and SogouCS news corpora pertaining to 5 categories, which are \textit{Sports}, \textit{Finance}, \textit{Entertainment}, \textit{Automobile} and \textit{Technology}. It contains 450,000 training samples and 60,000 samples for testing in total. Sogou New also has three columns in data set files, \emph{label}, \emph{title}, \emph{description}, similarly, we treat \emph{title} as abstract input and \emph{description} as contents input.

\subsection{Setting}

\paragraph{External knowledge matrix size} $128 \times 128$

\paragraph{Sequence length} we set the input sequence length as 256 (word level) in all experiments besides comparison experiments. 

\paragraph{Hyperparameters for classifiers} The number of LSTM cells is set to 128 for the three configurations. The self-attention mechanism we adopted is the single head self-attention with the attention width set to 16.

\section{Results and discussion}

\begin{table}[t]
\centering
\caption{Testing error of our model and related studies}
\label{tab:2}       
\begin{tabular}{llllll}
\hline\noalign{\smallskip}
Data set & AG & Sogou & AG(5k) & Sogou(10k)\\
\noalign{\smallskip}\hline\noalign{\smallskip}
\textit{- our model} \\
SeMemNN-ct & 9.29 & 4.73 & 16.72 & 10.82\\
SeMemNN-abs & 9.04 & 4.62 & 15.32& 9.80\\
B-SeMemNN-ct & 9.01 & 4.52 & 15.37 & 9.37\\
B-SeMemNN-abs & 8.68 & 4.19 & 14.35 & 8.76 \\
SAB-SeMemNN-ct & 8.88 & 4.33 & 14.07 & 7.95\\
SAB-SeMemNN-abs & \textcolor{red}{8.37} & \textcolor{red}{3.67} & \textcolor{red}{13.79} & \textcolor{red}{7.89}\\
\noalign{\smallskip}\hline\noalign{\smallskip}
\textit{- related studies} \\
Bow \cite{zhang2015character} & 11.19 & 7.15 & - & -\\
Bow TFIDF \cite{zhang2015character} & 10.36 & 6.55 & - & - \\
ngrams TFIDF \cite{zhang2015character} & 7.64 & 2.81 & - & - \\
Bag-of-means \cite{zhang2015character} & 16.91 & 10.79 & - & - \\
LSTM \cite{zhang2015character} & 13.94 & 4.82 & - & - \\
char-CNN \cite{zhang2015character} & 9.51 & 4.39 & - & -\\ 
VDCNN \cite{conneau2016very} & \textbf{8.67} & \textbf{3.18} & - & - \\
VDCNN & \textcolor{blue}{10.64} & \textcolor{blue}{6.53} & \textcolor{blue}{19.25} & \textcolor{blue}{unable} \\
XLNet \cite{yang2019xlnet} & 4.49 & - & - & - \\
ULMFiT \cite{rother2018ulmfit} & 5.01 & - & - & - \\
CNN \cite{johnson2014effective} & 6.57 & - & - & - \\
DPCNN \cite{johnson2017deep} & 6.87 & 1.84 & - & - \\
\noalign{\smallskip}\hline
\end{tabular}

\end{table}

\begin{figure*}[h!]
\centering
 
\subfigure[]{
    \begin{minipage}[t]{0.45\linewidth}
        \centering
        \includegraphics[width=1\textwidth]{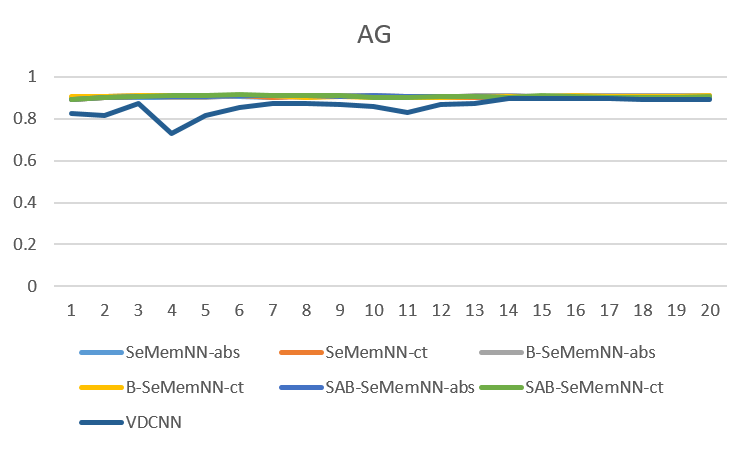}\\

    \end{minipage}
}
\subfigure[]{
    \begin{minipage}[t]{0.45\linewidth}
        \centering
        \includegraphics[width=1\textwidth]{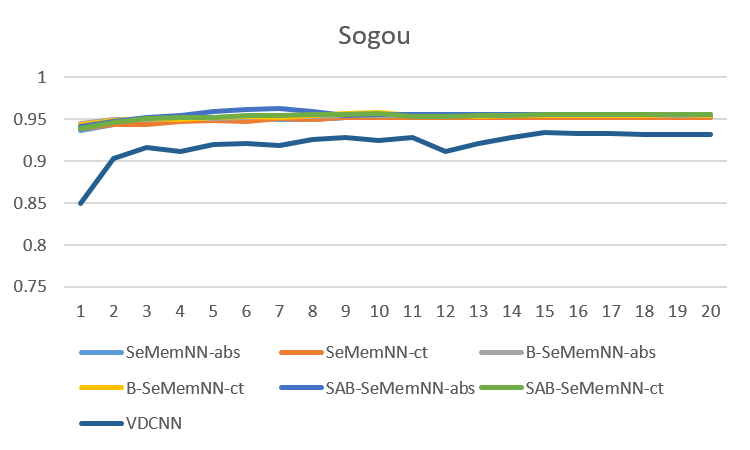}\\

    \end{minipage}
}
\subfigure[]{
    \begin{minipage}[t]{0.45\linewidth}
        \centering
        \includegraphics[width=1\textwidth]{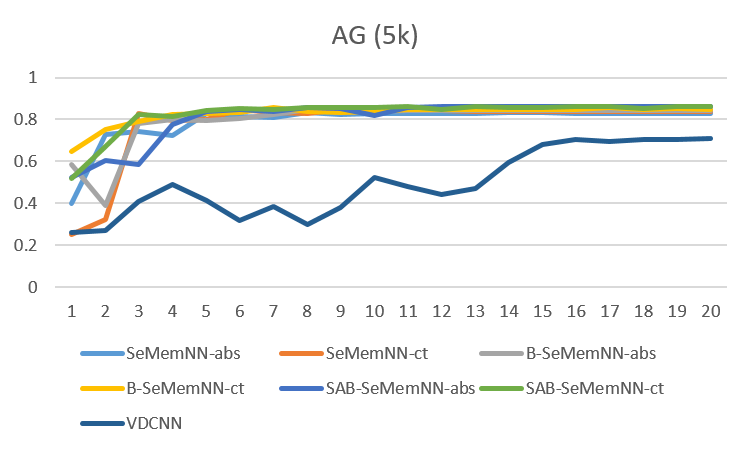}\\

    \end{minipage}
}
\subfigure[]{
    \begin{minipage}[t]{0.45\linewidth}
        \centering
        \includegraphics[width=1\textwidth]{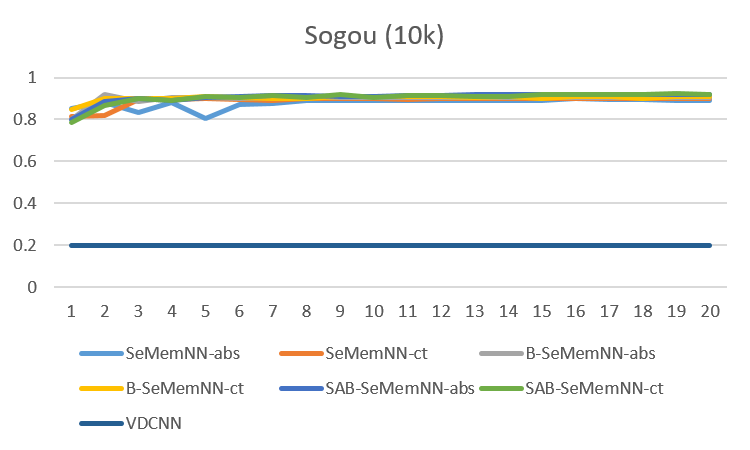}\\

    \end{minipage}
}
\caption{Validation curves}

\label{fig:compare_fig}
\end{figure*}

The experiment results are presented in Table 2. The best performances of our configurations are highlighted in red, which are 8.37, 3.67, 13.79 and 7.89 for the error rates of our proposed model on AG, Sogou, AG(5k), Sogou(10k) respectively. The bold numbers are the officially reported accuracy of VDCNN, to which our proposed model is close. The numbers in blue are the results coming from our comparison experiment by using VDCNN where we set the sequence length to 256 under word level. From these results we can see that our model outperforms VDCNN on AG News for the official error rate, and is very close to VDCNN's performance on Sogou News. If we set VCDNN with the same input sequence length (256) in word-level, the performance of our proposed model is obviously better. According to these results, we see the advantages of our model as follows:

\paragraph{Most of the contributions come from external matrix} From the results of Table 2, in the case of training with a large scale dataset, no matter we use simple LSTM or more complex bi-directional LSTM with self-attention, the testing error rates of different configurations are basically similar to each other. It can be said that such near state-of-the-art performance mainly attributes to the contribution from external memory. 

\paragraph{Using abstract to build the external memory is better than contents} Before the experiment, we assumed that using the description to construct the semantics matrix would be better than using the abstract because of the complete information the description contains. However, the testing results surprised us. From Table 2 we can see that the results of using the description to construct the semantics matrix are better than using the abstract. This result suggests that a good summary or title already contains the main information for judging which domain/topic the text belongs to. The most recent semantics used during the reading processing may come mostly from the abstracts. It might also imply that when humans memorize text contents, they would have a clearer memory of the key information rather than the whole contents. 

\paragraph{SeMemNN can still work on a few-shot learning} Table 2 shows that although we have greatly shrank the scale of the training set, our proposed method can still outperform VDCNN. After shrinking the scale of the data, the performance of VDCNN has been greatly decreased, especially for Sogou news, VDCNN has been unable to learn from the training samples. Table 3 presents some samples of these two data sets. We can see that the samples from Sogou News consists of Pinyin (the romanization of the Chinese characters based on their pronunciation in Mandarin), which may have more varieties of combinations than the samples from AG News. Perhaps because of such complex combinations, VDCNN can not work on Sogou News with few-shot learning.

\begin{table}[t]
\centering
\caption{Examples of samples}
\label{tab:3}       
\begin{tabular}{lp{6cm}}
\hline\noalign{\smallskip}
  Data set & Example \\
\noalign{\smallskip}\hline\noalign{\smallskip}
AG News &  `Dozens of Rwandan soldiers flew into Sudan's troubled Darfur region Sunday, ...'\\
Sogou News & `go1ng jia1o ta4i yo1ng ji3, chu1 zu1 che1 ta4i a2ng gui4, ma3i che1 ta4i ya2o yua3n ...' \\
\noalign{\smallskip}\hline
\end{tabular}
\end{table}

\paragraph{Comparing to the model with similar performance (VDCNN), our proposed can be trained faster}
Figure 2 shows the validation curves over the training period. It can be easily discovered that no matter whether the proposed method is trained with a large scale data set or few-shot learning, it learns faster and better than VDCNN.

\section{Conclusion and future works}
In this paper, we proposed a semantic matrix-based memory neural network (SeMemNN) for text classification. The proposed method was evaluated on two freely available large scale datasets and their sub-sets by shrinking the amount of training data. On one hand, the analysis shows that SeMemNN is an effective and time-efficient method of learning semantics comparing to the VDCNN. Regarding training with the large scale datasets, our model converges only after one iteration and achieves about 91\% and 96\% on AG News and Sogou News respectively. Regarding training with the small scale datasets, our model converges after five iterations and achieves about 85\% and 90\% on AG News and Sogou News respectively. On the other hand, we found that using abstracts to encode semantic matrix is better than using descriptions (full text). This result is very worthy for further study.

In future work, we will evaluate SeMemNN on other kinds of text classification datasets on a large and small scale. Additionally, we will try to figure out why using abstract to encode semantic matrix is better than using the full text.

\section*{Acknowledgment}
This work was partly supported by JST, ERATO, Grant Number JPMJER1401, and JST-Mirai Program Grant Number JPMJMI18C6.


\begin{thebibliography}{}
\bibitem{mitchell1980need}
Mitchell, T. M. (1980). The need for biases in learning generalizations (pp. 184-191). New Jersey: Department of Computer Science, Laboratory for Computer Science Research, Rutgers Univ..

\bibitem{joachims1998text}
Joachims, T. (1998, April). Text categorization with support vector machines: Learning with many relevant features. In European conference on machine learning (pp. 137-142). Springer, Berlin, Heidelberg.

\bibitem{lewis2004rcv1}
Lewis, D. D., Yang, Y., Rose, T. G., \& Li, F. (2004). Rcv1: A new benchmark collection for text categorization research. Journal of machine learning research, 5(Apr), 361-397.

\bibitem{johnson2014effective}
Johnson, R., \& Zhang, T. (2014). Effective use of word order for text categorization with convolutional neural networks. arXiv preprint arXiv:1412.1058.

\bibitem{zhang2015character}
Zhang, X., Zhao, J., \& LeCun, Y. (2015). Character-level convolutional networks for text classification. In Advances in neural information processing systems (pp. 649-657).

\bibitem{conneau2016very}
Conneau, A., Schwenk, H., Barrault, L., \& Lecun, Y. (2016). Very deep convolutional networks for text classification. arXiv preprint arXiv:1606.01781.

\bibitem{johnson2017deep}
Johnson, R., \& Zhang, T. (2017, July). Deep pyramid convolutional neural networks for text categorization. In Proceedings of the 55th Annual Meeting of the Association for Computational Linguistics (Volume 1: Long Papers) (pp. 562-570).

\bibitem{conneau2016very}
Conneau, A., Schwenk, H., Barrault, L., \& Lecun, Y. (2016). Very deep convolutional networks for text classification. arXiv preprint arXiv:1606.01781.

\bibitem{vilalta2002perspective}
Vilalta, R., \& Drissi, Y. (2002). A perspective view and survey of meta-learning. Artificial intelligence review, 18(2), 77-95.

\bibitem{weston2014memory}
Weston, J., Chopra, S., \& Bordes, A. (2014). Memory networks. arXiv preprint arXiv:1410.3916.

\bibitem{santoro2016one}
Santoro, A., Bartunov, S., Botvinick, M., Wierstra, D., \& Lillicrap, T. (2016). One-shot learning with memory-augmented neural networks. arXiv preprint arXiv:1605.06065.

\bibitem{yang2019xlnet}
Yang, Z., Dai, Z., Yang, Y., Carbonell, J., Salakhutdinov, R. R., \& Le, Q. V. (2019). Xlnet: Generalized autoregressive pretraining for language understanding. In Advances in neural information processing systems (pp. 5754-5764).

\bibitem{devlin2018bert}
Devlin, J., Chang, M. W., Lee, K., \& Toutanova, K. (2018). Bert: Pre-training of deep bidirectional transformers for language understanding. arXiv preprint arXiv:1810.04805.


\bibitem{sukhbaatar2015end}
Sukhbaatar, S., Weston, J., \& Fergus, R. (2015). End-to-end memory networks. In Advances in neural information processing systems (pp. 2440-2448).

\bibitem{miller2016key}
Miller, A., Fisch, A., Dodge, J., Karimi, A. H., Bordes, A., \& Weston, J. (2016). Key-value memory networks for directly reading documents. arXiv preprint arXiv:1606.03126.

\bibitem{lee2015high}
Lee, J., \& Tashev, I. (2015). High-level feature representation using recurrent neural network for speech emotion recognition. In Sixteenth annual conference of the international speech communication association.

\bibitem{le2017discretized}
Le, D., Aldeneh, Z., \& Provost, E. M. (2017, August). Discretized Continuous Speech Emotion Recognition with Multi-Task Deep Recurrent Neural Network. In Interspeech (pp. 1108-1112).

\bibitem{zhang2019attention}
Zhang, Z., Wu, B., \& Schuller, B. (2019, May). Attention-augmented end-to-end multi-task learning for emotion prediction from speech. In ICASSP 2019-2019 IEEE International Conference on Acoustics, Speech and Signal Processing (ICASSP) (pp. 6705-6709). IEEE.

\bibitem{zhou2016attention}
Zhou, P., Shi, W., Tian, J., Qi, Z., Li, B., Hao, H., \& Xu, B. (2016, August). Attention-based bidirectional long short-term memory networks for relation classification. In Proceedings of the 54th annual meeting of the association for computational linguistics (volume 2: Short papers) (pp. 207-212).

\bibitem{rother2018ulmfit}
Rother, K., \& Rettberg, A. (2018). Ulmfit at germeval-2018: A deep neural language model for the classification of hate speech in german tweets.



\end{thebibliography}
\end{document}